\documentclass[letterpaper,10pt,conference]{ieeeconf}

\IEEEoverridecommandlockouts
\usepackage{amsmath,amssymb,amsfonts}
\usepackage{bm}
\usepackage{booktabs}
\usepackage{tabularray}
\UseTblrLibrary{booktabs}
\usepackage{array}
\usepackage{multirow}
\usepackage{graphicx}
\usepackage{xcolor}
\usepackage{cite}
\usepackage{url}

\makeatletter
\let\NAT@parse\undefined
\makeatother
\usepackage[linkcolor=red,citecolor=red,urlcolor=blue,colorlinks=true]{hyperref}

\newcommand{\AlignedPair}[3][1.7em]{\makebox[#1][r]{#2}~/~\makebox[1.7em][r]{#3}}
\newcommand{\MethodHeader}[1]{\makebox[1.1cm][c]{\raisebox{-.5\height}{\shortstack[c]{#1}}}}
\newcommand{\SEthree}{\mathrm{SE}(3)}
\newcommand{\SOthree}{\mathrm{SO}(3)}

\title{
  NeuRIO: A Streaming Neural Estimator for \\
  Zero-Shot Sim-to-Real Multi-Robot Relative Inertial Odometry
}

\author{
    Zhehan Li\textsuperscript{1,2},
    Jiadong Lu\textsuperscript{1,2},
    Shengwei Ren\textsuperscript{3},
    Chao Xu\textsuperscript{1,2},
    and
    Yanjun Cao\textsuperscript{1,2}
    \thanks{
        \textsuperscript{1} State Key Laboratory of Industrial Control Technology,
        Institute of Cyber-Systems and Control,
        Zhejiang University,
        Hangzhou, China.
    }
    \thanks{
        \textsuperscript{2} Huzhou Institute of Zhejiang University,
        Huzhou, China.
    }
    \thanks{
        \textsuperscript{3} Hangzhou Guixing Intelligent Technology Co., Ltd.,
        Hangzhou, China.
    }
    \thanks{
        This work was supported by National Nature Science Foundation of China under Grant 62103368.
        The corresponding author is Yanjun Cao.
    }
    \thanks{
        E-mails: \texttt{zhehanli@zju.edu.cn}, \texttt{yanjuncao@zju.edu.cn}.
    }
}

\begin{document}

\maketitle

\begin{abstract}
  We present NeuRIO, a streaming neural estimator for anchor-free 6-DoF relative inertial odometry using only identified inter-robot bearings, ranges, and IMU measurements.
  NeuRIO canonicalizes measurements into gravity-aligned coordinates, represents robots as nodes and mutual observations as factors, and uses attention for spatial reasoning and GRUs for temporal modeling.
  As a graph network, NeuRIO applies shared node-wise and factor-wise operators throughout the network, enabling it to handle different team sizes and time-varying observation graphs.
  NeuRIO is trained on a simulator that couples various motion patterns, device-level sensor characteristics, and diverse, realistic modeled, and temporally persistent sensor corruptions.
  In this way, NeuRIO achieves zero-shot sim-to-real transfer.
  Across $24$ real-world sequences, NeuRIO achieves $14.1\,\mathrm{cm}$ position RMSE and $3.9^\circ$ rotation RMSE.
  More importantly, NeuRIO demonstrates strong computational scalability, maintaining an update cost below $20\,\mathrm{ms}$ with up to $400$ robots in simulation, while optimization-based methods exceed $20\,\mathrm{ms}$ at only $24$ robots.
  Moreover, even trained on limited team sizes, NeuRIO transfers directly to unseen larger teams without architectural or parameter changes.
\end{abstract}

\section{Introduction}
\label{sec:introduction}

Formation control, collision avoidance, and cooperative perception depend on accurate relative localization, especially when GNSS or a shared map is unavailable.
Camera bearings and UWB ranges are common choices for inter-robot sensing, and IMUs provide high-rate motion information.
Based on these measurements, filter and optimization methods \cite{xun2023crepes,li2025crepesx,lu2026ctrio} demonstrate accurate relative localization, while learned methods \cite{wang2025virgil,wang2026anyamber} show potential for direct regression of relative poses from bearing and range measurements.
However, they still rely on geometric refinement or pose priors at inference.
This leaves open whether a fully integrated end-to-end model can achieve accurate multi-robot relative localization, directly mapping inertial and inter-robot observations to 6-DoF relative poses without iterative optimization.

NeuRIO addresses this with a streaming neural estimator.
Firstly, it transforms inertial and bearing measurements into gravity-aligned coordinates, simplifying the input distribution and improving generalization across different mounting configurations.
Secondly, for spatial reasoning, it models robots as nodes and directed mutual observations as factors and performs node-to-factor and factor-to-node updates through cross-attention.
Such a graph architecture learns shared node-wise and factor-wise processing and information-fusion operators that are independent of graph size, enabling the same model to generalize to graphs of arbitrary size in principle.
Thirdly, for temporal modeling, it introduces recurrent states for nodes and factors in each layer, allowing the network to utilize temporal context.
The resulting estimator directly predicts relative poses from the measurement stream and recurrent state without iterative optimization during inference.

\begin{figure}[t]
  \centering
  \includegraphics[width=\columnwidth]{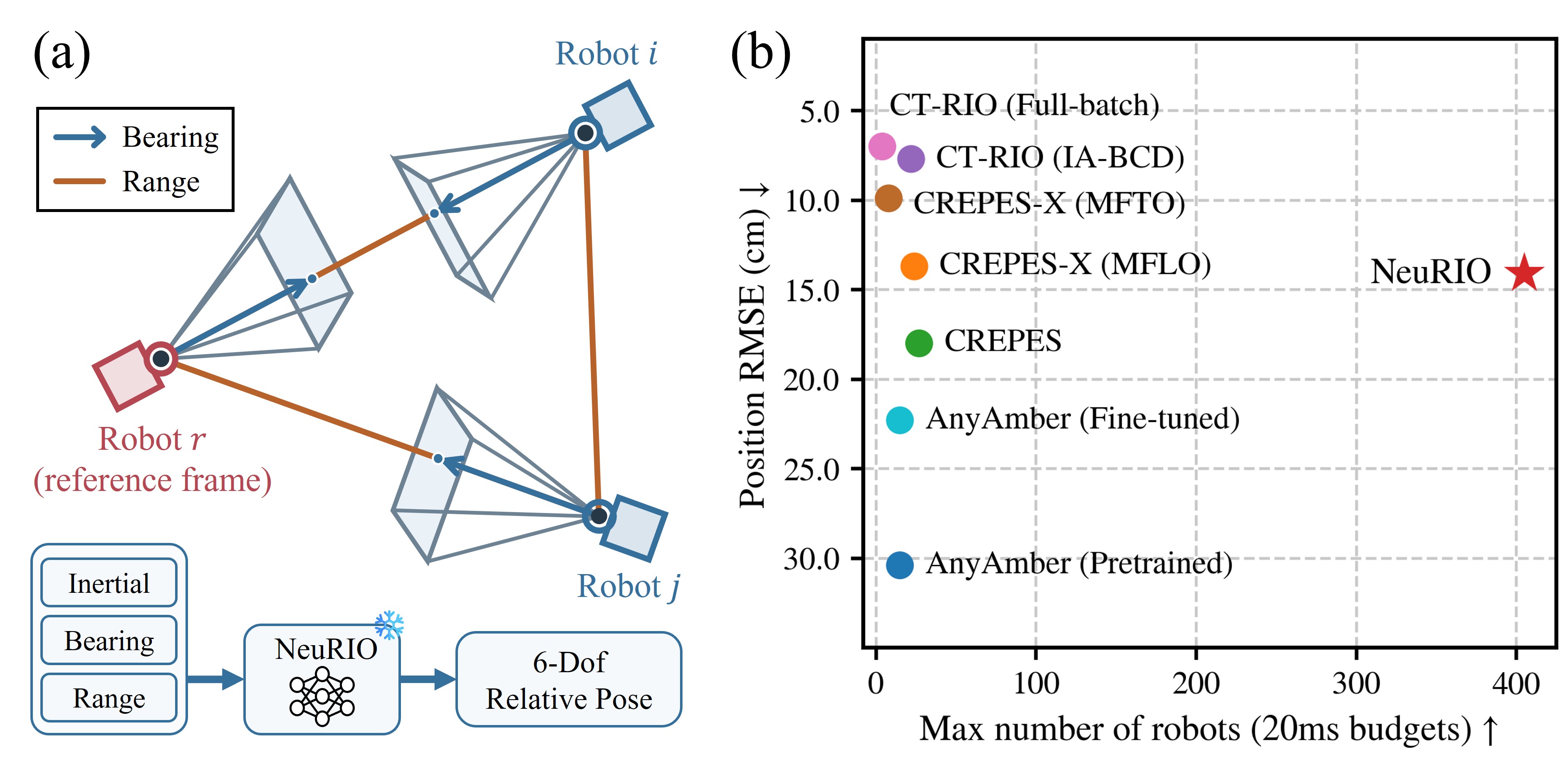}
  \caption{
    (a) Relative localization estimates neighboring robots' 6-DoF poses in a moving reference robot's frame from bearings, ranges, and inertial measurements.
    (b) Overall position RMSE versus the number of robots within $20\,\mathrm{ms}$ ($50\,\mathrm{Hz}$ camera frames).
    See detailed configurations in \autoref{sec:setup} (AnyAmber \cite{wang2026anyamber} excludes its MatchNet, and uses ground-truth as priors).
  }
  \label{fig:teaser}
\end{figure}

To support zero-shot sim-to-real transfer, we construct a simulator that models realistic sensor characteristics and failure processes.
We randomize both robot motion patterns and persistent device-level sensor characteristics.
Rather than treating sensor corruption as independent frame-wise noise, we explicitly model failures as temporally persistent processes.
This encourages the estimator to learn how sensing reliability evolves over time instead of relying only on instantaneous observations.
For bearing measurements, we model measurement noise, missing detections, flickering, blackouts, outliers, and identification errors.
For range measurements, we model measurement noise, device-specific bias, non-line-of-sight and multipath-induced outliers, and dropouts.
For IMU measurements, we account for measurement noise, sensor bias, axis misalignment, and attitude errors.
This structured simulation enables the frozen model to generalize directly to real-world sequences without real-world training, fine-tuning, or adaptation.

Across $24$ real-world sequences, the frozen model achieves position and rotation RMSEs of $14.1\,\mathrm{cm}$ and $3.9^\circ$, respectively.
The most closely related learning method, AnyAmber \cite{wang2026anyamber}, fails on $12$ of the $24$ sequences, and yields higher RMSE even with fine-tuning and ground-truth priors.
CREPES-X \cite{li2025crepesx} and CT-RIO \cite{lu2026ctrio} achieve lower RMSEs through iterative optimization, but at substantially higher computational cost.
In simulation test, their latencies exceed $20\,\mathrm{ms}$ at $24$ and $22$ robots, whereas NeuRIO remains within $20\,\mathrm{ms}$ through $400$.
This scalability is further supported by controlled team-size experiments, where models trained on smaller teams transfer directly to unseen larger teams in real world and much larger teams in simulation without architectural or parameter changes.

The contributions of this work are summarized as follows:
\begin{itemize}
  \item
        We formulate anchor-free multi-robot relative localization as streaming state estimation over a time-varying directed measurement graph, and introduce NeuRIO, a fully integrated end-to-end model that directly predicts 6-DoF relative poses without iterative optimization.
  \item
        We introduce a recurrent factor-graph architecture with gravity-aligned measurements and shared node and factor operators, enabling a single model to stream across changing observation graphs and team sizes without architectural or parameter changes.
  \item
        We develop a sim-to-real training simulator that models realistic sensor characteristics and failure processes, and point out that temporal persistence, missing observations, and outliers are key drivers of zero-shot transfer.
\end{itemize}
All source code and data will be released \footnote{\url{https://github.com/FAST-FIRE/NeuRIO}}.

\section{Related Works}
\label{sec:related}

\subsection{Relative Localization}

Relative localization estimates inter-robot states without requiring all robots to maintain accurate poses in a shared global frame.
Range-only \cite{cossette2021relative,fishberg2024murp} and bearing-only \cite{faessler2014monocular,stegagno2016ground} approaches are widely adopted in practice because of their sensing characteristics.
Combining the two modalities can exploit complementary geometric information and improve robustness to failures of individual sensing modalities.
The theoretical conditions for relative localizability under distance, angle, and self-displacement measurements have been studied in \cite{chen2025localizability}.
Practical systems have integrated camera bearings, UWB ranging, and IMU measurements \cite{xun2023crepes,li2025crepesx,lu2026ctrio} in filtering and optimization frameworks.
In particular, CREPES-X \cite{li2025crepesx} and CT-RIO \cite{lu2026ctrio} achieve high-accuracy relative pose estimation, but both rely on iterative optimization whose computational cost grows rapidly with problem size.
NeuRIO instead amortizes this process into a single forward pass, enabling real-time inference for large robot teams.

\subsection{Learning-based Relative Localization}

Multi-robot systems can be naturally represented as graphs.
Graph learning has progressed from message passing and neighborhood attention \cite{gilmer2017neural,velickovic2018graph} to graph transformers that encode nodes, edges, and graph structure directly in attention \cite{ying2021graphormer,dwivedi2021generalization}.
Related multi-robot perception systems use graph-based spatial encoding or cross-attention to fuse distributed visual observations \cite{zhou2022collaborative}.
More recent methods directly regress multi-robot poses from images.
CoViS-Net \cite{blumenkamp2025covisnet} estimates relative poses and local spatial context from visual observations, while Implicit Virtual Leader \cite{yang2026ivl} predicts 6-DoF poses in a learned formation frame using a Transformer-based graph neural network.
Besides images, learned graph models have been applied to UWB ranging and inertial odometry.
Neural Ranging Inertial Odometry \cite{wang2025neuralranging} combines recurrent inertial features with graph attention over UWB anchors and tags, but assumes fixed infrastructure, known anchor coordinates, and relies on geometric optimization.
Mr. Virgil \cite{wang2025virgil} uses a graph neural network to associate anonymous bearings with UWB ranges and predict initial positions and uncertainties, which are subsequently refined by differentiable pose graph optimization.
Building on Mr. Virgil, AnyAmber \cite{wang2026anyamber} introduces a heterogeneous EGAT architecture and formulates a generalist neural network that accommodates diverse localization settings, including anchor configurations, UWB tag layouts, and the presence or absence of bearing observations.
Yet, AnyAmber \cite{wang2026anyamber} still relies on pose graph optimization at inference, and its performance degrades when the optimization is disabled.
In contrast, NeuRIO focuses on a fully network driven estimator that directly predicts relative poses without optimization.

\subsection{Sim-to-Real Transfer}

The performance of neural networks depends strongly on the coverage and diversity of their training data, while collecting real-world data is often costly and time-consuming, especially for multi-robot systems.
Simulation provides an alternative, allowing training data to be generated under controlled variations with accurate supervision.

Domain randomization exploits this flexibility by varying task-relevant properties of the simulated data to improve robustness to real-world distribution shifts.
In robotics, visual randomization has enabled transfer for object localization and vision-based aerial navigation \cite{tobin2017domain,sadeghi2016cad2rl}, while randomization of dynamics, actuator properties, sensor noise, latency, and environmental conditions has been widely used for manipulation and legged locomotion \cite{peng2018sim,tan2018sim}.
These results highlight that sim-to-real performance depends not only on simulation fidelity, but also on how the training distribution represents the variations expected during deployment.
Following this principle, NeuRIO randomizes robot motion patterns, device-level sensor characteristics, and temporally persistent sensing failures to improve zero-shot transfer to real-world multi-robot relative localization.

\section{Problem Formulation}
\label{sec:problem}

\subsection{Multi-Robot State and Mutual Observation}

Consider one streaming sequence with a fixed set of $n$ robots indexed by $\mathcal{N}=\{0,\ldots,n-1\}$ and a designated reference robot $r$.
At time $t$, robot $i$ has pose ${}^{W}\mathbf{T}_{R_i}^{t}=({}^{W}\mathbf{R}_{R_i}^{t},{}^{W}\mathbf{p}_{R_i}^{t})\in\SEthree$ in a global world frame.
The relative transform of robot $j$ in robot $i$ is
\begin{align}
  {}^{R_i}\mathbf{R}_{R_j}^{t}
   & = ({}^{W}\mathbf{R}_{R_i}^{t})^{-1}{}^{W}\mathbf{R}_{R_j}^{t},
  \\
  {}^{R_i}\mathbf{p}_{R_j}^{t}
   & = ({}^{W}\mathbf{R}_{R_i}^{t})^{-1}({}^{W}\mathbf{p}_{R_j}^{t}-{}^{W}\mathbf{p}_{R_i}^{t}).
  \label{eq:relative_pose}
\end{align}

As shown in \autoref{fig:teaser} (a), for an ordered pair $(i,j)$, the calibrated camera of robot $i$ produces a unit bearing ${\hat{\mathbf{z}}_b}{}^{t}_{R_i \to R_j}\in\mathbb{S}^{2}$ associated with robot $j$, and the UWB produces a range ${\hat{z}_d}{}^{t}_{R_i \to R_j} \in \mathbb{R}^{1}$.
The nominal measurement models are
\begin{equation}
  {\hat{\mathbf{z}}_b}{}^{t}_{R_i \to R_j}
  = \frac{{}^{R_i}\mathbf{p}_{R_j}^{t}}{\|{}^{R_i}\mathbf{p}_{R_j}^{t}\|_2},
  \qquad
  {\hat{z}_d}{}^{t}_{R_i \to R_j}
  = \|{}^{R_i}\mathbf{p}_{R_j}^{t}\|_2.
  \label{eq:measurement_model}
\end{equation}
An available measurement may contain false detection, identity switch, and outlier.
From $t-1$ to $t$, robot $i$ supplies the high-rate inertial window $\{({\hat{\mathbf{z}}_i}{}^{\tau}_{R_i},\Delta\tau)\}$.
Gravity measurement ${\hat{\mathbf{z}}_g}{}^{t}_{R_i}$ is provided by complementary filter \cite{mahony2008nonlinear}.

\subsection{Streaming Relative State Estimation}

At each time $t$, the available inter-robot measurements define a directed observation graph $\mathcal{G}^{t}=(\mathcal{N},\mathcal{E}^{t})$, where an edge $(i,j)\in\mathcal{E}^{t}$ represents the measurements of robot $i$ associated with robot $j$.
Each measurement is accompanied by a binary mask $m$ that records whether a value is present, but does not assert that it is correct.
The graph is directed because the observations are expressed in the observer's local frame.
Let $\mathcal{I}^{t}_{R_i}$ denote the inertial measurements of robot $i$ between two consecutive estimation steps.

The objective is to estimate relative pose ${}^{R_r}\widehat{\mathbf{T}}_{R_i}^{t}$ of other robots $i \in \mathcal{N} \setminus \{r\}$ in the reference frame using only measurements available up to the current time.
The input of the network at time $t$ contains all $\mathcal{G}^{\tau}$ from $\tau=0$ to $t$ and all $\mathcal{I}^{\tau}_{R_i}$ from $\tau=1$ to $t$ of robot $i \in \mathcal{N}$ (as $\mathcal{I}^{0}$ cannot be defined).
We formulate the estimator $f_{\theta}$ as
\begin{equation}
  \{{}^{R_r}\widehat{\mathbf{T}}_{R_i}^{t}\}_{i \in \mathcal{N} \setminus \{r\}}
  =
  f_{\theta}
  \left(
  \{\mathcal{G}^{\tau}\}_{\tau=0}^{t},
  \{\mathcal{I}^{\tau}_{R_i}\}_{\tau=1, i \in \mathcal{N}}^{t}
  \right).
  \label{eq:streaming_objective}
\end{equation}

We aim to design such an estimator, which should remain applicable to different numbers and compositions of robots, tolerate missing and corrupted inter-robot observations, and operate online in a streaming manner.

\section{Network Architecture}
\label{sec:method}

\begin{figure}[t]
  \centering
  \includegraphics[width=.98\columnwidth]{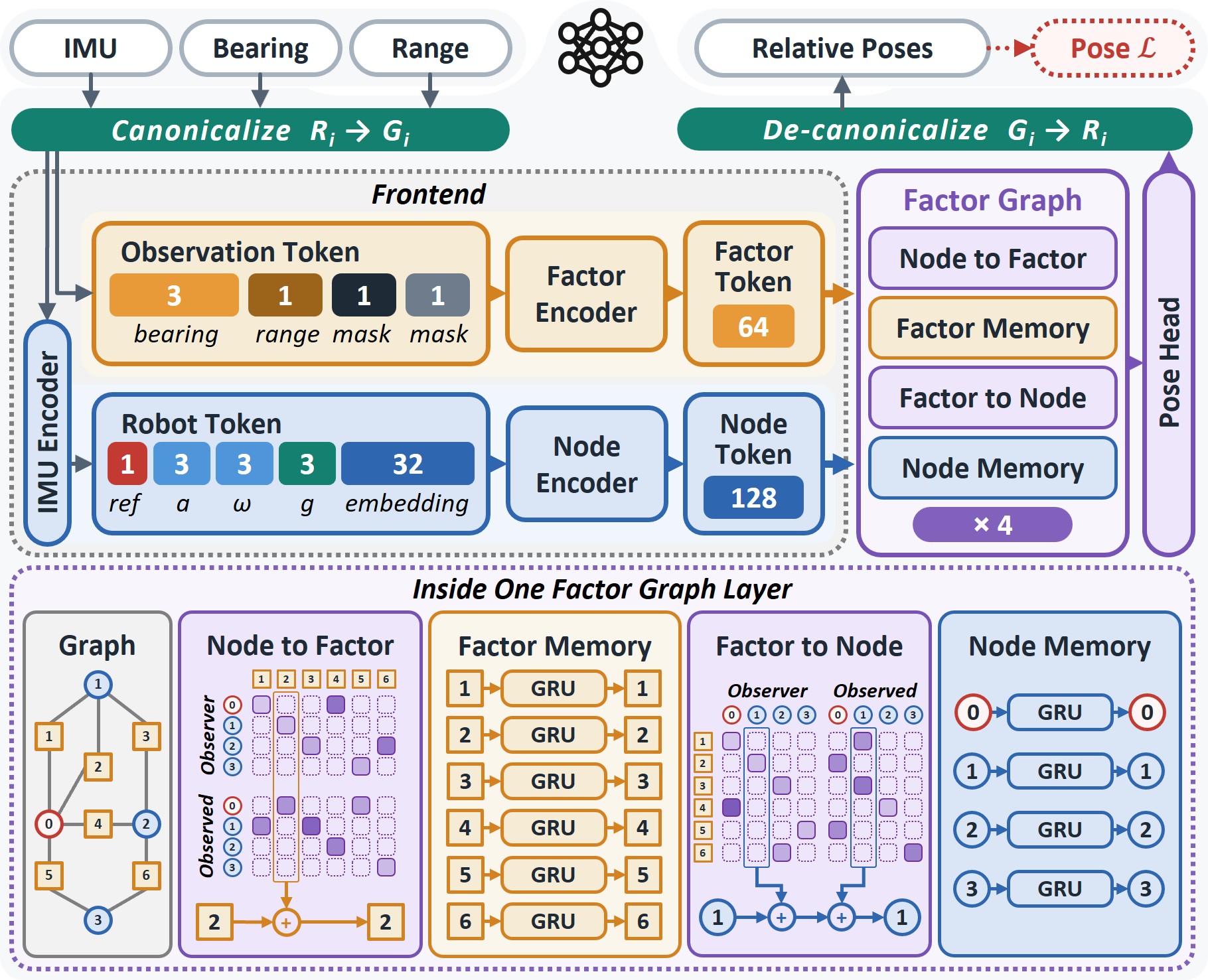}
  \caption{
    Architecture of NeuRIO, which directly predicts relative poses from bearings, ranges, and IMU measurements.
  }
  \label{fig:overview}
\end{figure}

\subsection{Overview}

As shown in \autoref{fig:overview}, NeuRIO first applies gravity-aligned canonicalization to express the measurements in a geometrically consistent representation.
This can reduce the impact of roll- and pitch-related degrees of freedom on the input data distribution, simplifying the problem the network needs to learn \cite{yan2018ridi,herath2020ronin,liu2020tlio}.
A frontend then encodes the raw observation inputs into node and factor tokens, which are processed by recurrent factor graph layers.
Within each layer, robot nodes and their associated observation factors exchange information through attention while preserving the asymmetric roles of the observing and observed robots.
Each layer maintains recurrent state for both node and factor tokens, providing temporal context throughout the spatial reasoning process.
Finally, a shared decoder maps the final-layer node representations to relative poses and associated uncertainties, which are supervised by the ground-truth relative poses.

\subsection{Gravity-Aligned Canonicalization}
\label{sec:canonical}

Let $^{R_i}{\hat{\mathbf{z}}_g}^{t}$ be the gravity direction reported in robot $i$'s body frame.
We define the gravity-aligned canonical frame $G_i$ through the minimal-tilt rotation ${}^{R_i}\mathbf{R}_{G_i}^{t}\in\SOthree$ that satisfies $({}^{R_i}\mathbf{R}_{G_i}^{t})^{-1}{}^{R_i}{\hat{\mathbf{z}}_g}{}^{t}=(0,0,-1)$, without introducing an arbitrary heading.
Every body-frame vector ${}^{R_i}\mathbf{v}^{t}$ is expressed in this canonical frame as ${}^{G_i}\mathbf{v}^{t}=({}^{R_i}\mathbf{R}_{G_i}^{t})^{-1}\,{}^{R_i}\mathbf{v}^{t}$.
The network predicts the relative pose $({}^{G_r}\mathbf{R}_{G_i}^{t},{}^{G_r}\mathbf{p}_{G_i}^{t})$ in this canonical frame, and the relative pose in the original body frame $({}^{R_r}\mathbf{R}_{R_i}^{t},{}^{R_r}\mathbf{p}_{R_i}^{t})$ can be recovered through
\begin{align}
  {}^{R_r}\widehat{\mathbf{R}}_{R_i}^{t}
   & = {}^{R_r}\mathbf{R}_{G_r}^{t}{}^{G_r}\widehat{\mathbf{R}}_{G_i}^{t}({}^{R_i}\mathbf{R}_{G_i}^{t})^{-1},
  \\
  {}^{R_r}\widehat{\mathbf{p}}_{R_i}^{t}
   & = {}^{R_r}\mathbf{R}_{G_r}^{t}{}^{G_r}\widehat{\mathbf{p}}_{R_i}^{t}={}^{R_r}\mathbf{R}_{G_r}^{t}{}^{G_r}\widehat{\mathbf{p}}_{G_i}^{t}.
  \label{eq:canonical_decode}
\end{align}

After canonicalization, ${}^{G_i}{\hat{\mathbf{z}}_g}^{t}=(0,0,-1)$ is constant.
We retain ${}^{R_i}{\hat{\mathbf{z}}_g}{}^{t}$ in the input vector to preserve the original attitude information.

\subsection{Frontend Encoding}

The robot and observation input is formulated as
\begin{equation}
  \begin{aligned}
    \mathbf{x}_{i,t}^{\mathrm{rob}}
     & = \left[
           b_i^{\mathrm{ref}},
             {}^{G_i}\bar{\mathbf{a}}_{R_i}^{t},
           {}^{G_i}\bm{\omega}_{R_i}^{t},
           {\hat{\mathbf{z}}_g}{}^{t}_{R_i},
           ~\psi_I(\mathcal{I}_{R_i}^{t})
           \right],             \\
    \mathbf{x}_{ij,t}^{\mathrm{obs}}
     & = \left[
           {}^{G_i}{\hat{\mathbf{z}}_b}{}^{t}_{R_i \to R_j},
           {\bar{z}_d}{}^{t}_{R_i \to R_j},
           m_b{}^{t}_{R_i \to R_j},
           m_d{}^{t}_{R_i \to R_j}
           \right],
  \end{aligned}
  \label{eq:node_input}
\end{equation}
where $b_i^{\mathrm{ref}}$ is a binary reference flag, $\psi_I$ is a GRU that maps the gravity-aligned inter-frame sequence of acceleration and angular rate to a 32-dimensional feature, $m_b{}^{t}_{R_i \to R_j}$ and $m_d{}^{t}_{R_i \to R_j}$ are binary masks indicating the availability of bearing and range measurements.
Acceleration ${}^{G_i}\bar{\mathbf{a}}_{i,t} = {}^{G_i}{\mathbf{a}}_{i,t} / g$ is normalized by the standard gravitational acceleration $g=9.81\,\mathrm{m/s^2}$, and range  ${\bar{z}_d}{}^{t}_{R_i \to R_j}={\hat{z}_d}{}^{t}_{R_i \to R_j} / d$ is normalized by a characteristic distance of $d=10\,\mathrm{m}$.

The node and factor encoders are conducted by two linear layers with GELU activation, and followed by a GRU to accumulate inertial and directed-observation histories.
The node encoder maps $\mathbf{x}_{i,t}^{\mathrm{rob}}$ to $\mathbf{h}_{i,t}\in\mathbb{R}^{128}$, and the factor encoder maps $\mathbf{x}_{ij,t}^{\mathrm{obs}}$ to $\mathbf{e}_{ij,t}\in\mathbb{R}^{64}$.

\subsection{Recurrent Factor Graph Layers}
\label{sec:backbone}

Each layer alternates spatial aggregation and temporal integration, first updating the factors and then the robot nodes.
Let $\mathbf{h}_{i,t}^{(\ell)}$ and $\mathbf{e}_{ij,t}^{(\ell)}$ denote the robot and factor states entering layer $\ell$.
$\bar{\mathbf{h}}_{i,t}^{(\ell)}$ and $\bar{\mathbf{e}}_{ij,t}^{(\ell)}$ denote the corresponding layer-normalized states.
Dropout is omitted for readability.
We denote query, key, value projections by $Q$, $K$, $V$, with subscripts distinguishing factor ($e$) and node ($n$) updates, and superscripts distinguishing layers $(\ell)$.

A bearing measurement is directed: robot $i$ observes robot $j$, but not necessarily vice versa.
Accordingly, the corresponding edge is directed, and the network needs to distinguish the two endpoint roles.
For $\mathbf{e}_{ij,t}^{(\ell)}$, we designate robot $i$ as the observer ($\mathrm{o}$) and robot $j$ as the observed ($\mathrm{s}$).
The attention scores are computed as
\begin{equation}
  a_{ij,t,\rho}^{(\ell)} = \frac{Q_{\rho}^{(\ell)}(\bar{\mathbf{h}}_{u_\rho,t}^{(\ell)})^{\top}
    K^{(\ell)}(\bar{\mathbf{e}}_{ij,t}^{(\ell)})}{\sqrt{d_h}} + b_{\rho}^{(\ell)}.
  \label{eq:edge_attention}
\end{equation}
Here $\rho\in\{\mathrm{o},\mathrm{s}\}$, with $u_{\mathrm{o}}=i$ and $u_{\mathrm{s}}=j$.
We denote the corresponding normalized attention weights by $\alpha_{ij,t,\rho}^{(\ell)}$, obtained by applying a softmax over the available factors incident to each robot under the corresponding endpoint role.
A directed factor is available when either its bearing mask or its range mask is valid.
Each directed factor interacts only with its two endpoint nodes, yielding an attention computation cost of $\mathcal{O}(|\mathcal{E}_t|)$, which becomes $\mathcal{O}(n^2)$ only for a dense all-pairs observation graph.

\subsubsection{Spatial Factor Update}
Each available factor first aggregates the current states of its two endpoint robots through independent sigmoid gates $\sigma$ of the attention scores.
The two contributions are concatenated and fused,
\begin{equation}
  \begin{aligned}
    \widetilde{\mathbf{e}}_{ij,t}^{(\ell)}
      & = \mathbf{e}_{ij,t}^{(\ell)}
    \\
    + & \mathrm{MLP}_{e}^{(\ell)} ([\sigma(a_{ij,t,\mathrm{o}}^{(\ell)})V_{e,\mathrm{o}}^{(\ell)}(\bar{\mathbf{h}}_{i,t}^{(\ell)}),\sigma(a_{ij,t,\mathrm{s}}^{(\ell)})V_{e,\mathrm{s}}^{(\ell)}(\bar{\mathbf{h}}_{j,t}^{(\ell)})]).
  \end{aligned}
  \label{eq:edge_update}
\end{equation}
The resulting $\widetilde{\mathbf{e}}_{ij,t}^{(\ell)}$ incorporates current endpoint information and provides the input to temporal integration.

\subsubsection{Temporal Factor Update}
Each factor then integrates its history through a GRU.
The resulting $\mathbf{e}_{ij,t}^{(\ell+1)}$ supplies the values for node aggregation.
If the measurement is unavailable at time $t$, the corresponding recurrent state $\mathbf{c}_{ij,t}^{(\ell)}$ is held until it becomes available again.

\subsubsection{Spatial Node Update}

Each robot node next aggregates the updated factors related to it.
Both the observer and observed-robot roles are considered, with independent attention scores and value projections for each role, to account for the asymmetric nature of directed observations.
Using the normalized attention weights, the two aggregates are fused,
\begin{equation}
  \begin{aligned}
    \widetilde{\mathbf{h}}_{i,t}^{(\ell)} = \mathbf{h}_{i,t}^{(\ell)}
     & + \mathrm{MLP}_{n,\mathrm{o}}^{(\ell)} (\sum_j \alpha_{ij,t,\mathrm{o}}^{(\ell)}V_{n,\mathrm{o}}^{(\ell)}(\bar{\mathbf{e}}_{ij,t}^{(\ell+1)}))
    \\
     & + \mathrm{MLP}_{n,\mathrm{s}}^{(\ell)} (\sum_j \alpha_{ji,t,\mathrm{s}}^{(\ell)}V_{n,\mathrm{s}}^{(\ell)}(\bar{\mathbf{e}}_{ji,t}^{(\ell+1)})).
  \end{aligned}
  \label{eq:factor_to_node}
\end{equation}
The resulting $\widetilde{\mathbf{h}}_{i,t}^{(\ell)}$ incorporates the available relational information and provides the input to temporal integration.

\subsubsection{Temporal Node Update}

Each node finally integrates its history through a GRU.
The resulting $\mathbf{h}_{i,t}^{(\ell+1)}$, together with $\mathbf{e}_{ij,t}^{(\ell+1)}$, forms the input to the next layer.

\subsection{Prediction and Supervision}
\label{sec:loss}

The shared pose head predicts a continuous 6D rotation representation \cite{zhou2019continuity}, a translation, and scalar position and rotation log variances for every robot.
The predicted poses in body-frame are recovered from \eqref{eq:canonical_decode}.
We supervise the relative poses using simulation ground-truth, excluding the reference robot $r$.
Following the heteroscedastic uncertainty-weighting approach of Kendall and Gal \cite{kendall2017uncertainties}, we construct position and rotation losses $\mathcal{L}_p$ and $\mathcal{L}_R$ from the squared Euclidean position error and the rotation error $d_R^2(\widehat{\mathbf{R}},\mathbf{R})=3-\operatorname{tr}(\widehat{\mathbf{R}}^{\top}\mathbf{R})$, respectively.
The total objective is averaged over robots $j\in\mathcal{N}\setminus\{r\}$ and frames in the training chunk:
\begin{equation}
  \mathcal{L}
  = \lambda_p\mathcal{L}_{p}
  + \lambda_R\mathcal{L}_{R},
  \qquad
  \lambda_p=1,\ \lambda_R=3.
  \label{eq:total_loss}
\end{equation}

The architecture is inherently agnostic to team size and reference-robot identity.
Its spatial computation operates directly on the measurement graph via local node-factor interactions, while temporal operators are shared across robots and directed robot pairs.
Thus, the same trained model can be deployed across different team sizes and reference robots without retraining or architectural and parameter changes.

\section{Training Simulator Design}
\label{sec:sim2real}

To support zero-shot sim-to-real deployment, we construct a simulator that models realistic sensor characteristics and failure processes.
The simulation parameters can be adjusted to cover diverse realistic conditions.
In this work, their values are determined from independent recordings collected on CREPES-X devices \cite{li2025crepesx}, without using the evaluation sequences.

\subsection{Environment and Motion Randomization}

The dataset contains $400$ sequences of $300\,\mathrm{s}$ with $3$ to $10$ devices.
Each sequence is generated in a cubic workspace with half-extent $h$ sampled from $1.5$ to $8.0\,\mathrm{m}$ and $0$ to $20$ spherical obstacles whose radii range from $0.01h$ to $0.10h$.
Obstacles are static with probability $0.50$; otherwise, they follow independent B-spline trajectories with control-knot rates from $0.2$ to $1.0\,\mathrm{Hz}$.
The obstacle determines both camera occlusion and UWB non-line-of-sight conditions.

Motion is conditioned on platform type.
Aerial devices follow three-dimensional B-spline trajectories with control-knot rates from $0.2$ to $1.0\,\mathrm{Hz}$ and speed commands from $0.5$ to $1.2\,\mathrm{m/s}$.
Their roll and pitch follow the simulated thrust direction, while yaw varies independently.
Ground devices follow planar differential-drive trajectories at $0.7$ to $1.6\,\mathrm{m/s}$, with occasional stops and varying turn rates.
Handheld devices combine smooth translational and rotational motions with slowly varying activity levels, producing transitions between quiet holding and sustained motion.
Static devices retain a randomized fixed pose throughout the sequence.
Together, these models expose the network to substantially different platform-dependent motion statistics.

\subsection{Sensor and Failure Randomization}

Each device receives a random mounting rotation that remains fixed throughout the sequence. 
Mounting rotations are sampled from two distributions: near-upright or arbitrary. 
Near-upright orientations use a yaw angle sampled uniformly from $0^\circ$ to $360^\circ$ and a small tilt angle sampled from a zero-mean Gaussian distribution with a standard deviation of $5^\circ$, while arbitrary orientations are sampled uniformly from the rotation group $\SOthree$.
Randomized sensor characteristics, dropout processes, and corruption processes are then applied to the nominal bearing, range, and IMU measurements, as summarized in \autoref{tab:simulator_sensor_models}. 
Persistent failures are modeled temporally rather than independently at each frame, allowing the simulator to reproduce both brief measurement losses and sustained sensor degradation.
This enables the network to learn to exploit temporal context and cross-robot redundancy to mitigate the effects of missing or corrupted measurements.

\begin{table}[!t]
  \centering
  \caption{
    Sensor configurations and failure processes.
  }
  \label{tab:simulator_sensor_models}
  \setlength{\tabcolsep}{2pt}
  \resizebox{\columnwidth}{!}{%
  \begin{tblr}{
    colspec={Q[c,m,wd=1.1em] Q[c,m,wd=1.1em] Q[l,m,wd=0.2\columnwidth] Q[l,m,wd=1.0\columnwidth]},
    colsep=\tabcolsep,
    column{1}={leftsep=0pt},
    column{4}={rightsep=0pt},
    rowsep=2pt,
  }
    \toprule
    \SetCell[r=10]{m} \rotatebox[origin=c]{90}{Bearing} & \SetCell[r=3]{m} \rotatebox[origin=c]{90}{Base}       &
    \textit{Noise} & Rotate about an isotropically sampled axis by a angle with $\sigma_0\sim U(0.1,4.0)^\circ$, multiplied by a device-specific scale $U(0.5,2.2)$. \\
    \cmidrule[l=-1]{3-4}
                                     &                                      &
    \textit{Field-of-view} & Assign omnidirectional cameras with probability $0.25$, otherwise a half-angle $U(60.0,115.0)^\circ$. Outside bearings are removed. \\
    \cmidrule[l=-1]{3-4}
    & &
    \textit{Occlusion} & Remove bearings whose connecting segment intersects an obstacle. \\
    \cmidrule[l=-1]{2-4}
                                     & \SetCell[r=5]{m} \rotatebox[origin=c]{90}{Dropout}    &
    \textit{Missed} & Independently miss up to $20\%$ of candidates. \\
    \cmidrule[l=-1]{3-4}
    & &
    \textit{Flicker} & Eligible links switch off with probability $\min(1,(1-p)/\kappa)$ and on with $\min(1,p/\kappa)$, where $\kappa\sim U(0.5,2.0)$ frames and $p \sim U(0.30,0.95)$. \\
    \cmidrule[l=-1]{3-4}
                                     &                                      &
    \textit{Short blackout} & Short outages have $\lambda\leq0.15\,\mathrm{s}^{-1}$ and $\tau$ from $0.5$ to $3.0\,\mathrm{s}$. \\
    \cmidrule[l=-1]{3-4}
    & &
    \textit{Long blackout} & Long outages have $\lambda\leq0.01\,\mathrm{s}^{-1}$ and $\tau$ from $8$ to $60\,\mathrm{s}$. \\
    \cmidrule[l=-1]{3-4}
                                     &                                      &
    \textit{Motion-related blackout} & Remove all bearings with $\tau=0.5\,\mathrm{s}$ and $\lambda=0.05s[1+2\min(m,1.5)]$, where $s\sim U(0,0.8)$ and $m=\|v\|/(1\,\mathrm{m/s})+\|\omega\|/(1\,\mathrm{rad/s})$. \\
    \cmidrule[l=-1]{2-4}
                                     & \SetCell[r=2]{m} \rotatebox[origin=c]{90}{Corruption} &
    \textit{False positive} & Replace bearing with probability $\min(0.5,ru)$, where $r=0$ with probability $0.50$, otherwise $r\sim U(0,0.03)$; device factor $u\sim U(0.4,1.6)$ is multiplied by $U(4,15)$ with probability $q\sim U(0,0.5)$; rotate the current bearing by $U(20.0,\theta_{\max})^\circ$ about a random perpendicular axis, with $\theta_{\max}\sim U(60.0,120.0)$. \\
    \cmidrule[l=-1]{3-4}
                                     &                                      &
    \textit{Identity switch} & Replace each bearing with its angularly nearest alternative among the observer's detections with probability $0.80$, retaining the original identity; $\tau=1.5\,\mathrm{s}$ and $\lambda=s\ell$, with $s\sim U(0,1.6)$ and $\ell\sim U(0,0.03)\,\mathrm{s}^{-1}$. \\
    \midrule
    \SetCell[r=6]{m} \rotatebox[origin=c]{90}{Range}    & \rotatebox[origin=c]{90}{Base}                                 &
    \textit{Line-of-sight noise} & Add zero-mean Gaussian noise with base $\sigma_0\sim U(2.0,12.0)\,\mathrm{cm}$, multiplied by a device-specific $U(0.6,1.7)$. \\
    \cmidrule[l=-1]{2-4}
                                     & \rotatebox[origin=c]{90}{Dropout}                              &
    \textit{Packet loss} & Mask each ordered link using a two-state process, with $\lambda\leq0.05\,\mathrm{s}^{-1}$ and $\tau$ from $0.02$ to $0.30\,\mathrm{s}$.                                                                                                                                                              \\
    \cmidrule[l=-1]{2-4}
                                     & \SetCell[r=4]{m} \rotatebox[origin=c]{90}{Corruption} &
    \textit{Antenna-delay residual} & Assign each device a fixed $b_i\in[-5.0,5.0]\,\mathrm{cm}$ and add $b_i+b_j$ to range $(i,j)$.                                                                                                                                                                                            \\
    \cmidrule[l=-1]{3-4}
                                     &                                      &
    \textit{Non-line-of-sight bias} & Add a positive exponential bias when the link is occluded, with mean up to $10.0\,\mathrm{cm}$; the same obstacles also cause camera occlusion.                                                                                                                                           \\
    \cmidrule[l=-1]{3-4}
                                     &                                      &
    \textit{Multipath outlier} & Add independent positive exponential errors with probability $U(0,0.02)$ and mean $U(10.0,50.0)\,\mathrm{cm}$. \\
    \cmidrule[l=-1]{3-4}
                                     &                                      &
    \textit{Degraded} & Select each device with probability $q\sim U(0,0.3)$; multiply both its multipath-outlier probability and exponential mean by a sequence-level factor $U(4,15)$ throughout the sequence. \\
    \midrule
    \SetCell[r=7]{m} \rotatebox[origin=c]{90}{IMU}    & \rotatebox[origin=c]{90}{Base}        &
    \textit{Noise} & Perturb accelerometer and gyroscope measurements with fixed per-axis standard deviations $0.1\,\mathrm{m/s^2}$ and $0.01\,\mathrm{rad/s}$, respectively.                                                                                                                                             \\
    \cmidrule[l=-1]{2-4}
                                     & \SetCell[r=5]{m} \rotatebox[origin=c]{90}{Corruption} &
    \textit{Turn-on bias} & Sample each axis bias from $\mathcal{N}(0,\sigma_b^2)$, with $\sigma_b\sim U(0.02,0.15)\,\mathrm{m/s^2}$ for the accelerometer and $\sigma_b\sim U(0.001,0.015)\,\mathrm{rad/s}$ for the gyroscope. \\
    \cmidrule[l=-1]{3-4}
                                     &                                      &
    \textit{Bias drift} & Accumulate independent Gaussian bias increments with per-axis standard deviations $10^{-3}s_b\Delta t\,\mathrm{m/s^2}$ and $10^{-4}s_b\Delta t\,\mathrm{rad/s}$, where $s_b\sim U(1,5)$ and $\Delta t=0.01\,\mathrm{s}$ is evaluated in seconds. \\
    \cmidrule[l=-1]{3-4}
    & &
    \textit{Scale-factor} & Sample each axis multiplicative scale from $\mathcal{N}(1,\sigma_s^2)$, clipped to $[0.8,1.2]$, with $\sigma_s\sim U(0.003,0.020)$ for the accelerometer and $\sigma_s\sim U(0.002,0.010)$ for the gyroscope. \\
    \cmidrule[l=-1]{3-4}
                                     &                                      &
    \textit{Axis misalignment} & Apply a device-fixed rotation about an isotropically sampled axis with angle $\theta\sim\mathcal{N}(0,\sigma_\theta^2)$, where $\sigma_\theta\sim U(0.2,1.5)^\circ$. \\
    \cmidrule[l=-1]{3-4}
    & &
    \textit{Vibration} & Add OU vibration with $(\sigma_a,\sigma_g;\tau_a,\tau_g)$ in $\mathrm{m/s^2}$, $\mathrm{rad/s}$, and seconds: aerial $(U(0.2,0.6),U(0.05,0.12);U(0.05,0.15),U(0.05,0.15))$, ground $(U(2.5,5.0),U(0.03,0.08);U(0.003,0.008),U(0.05,0.15))$, handheld $(U(0.03,0.15),U(0.01,0.04);U(0.01,0.04),U(0.02,0.08))$, and static $(U(0.01,0.05),U(0.004,0.015);U(0.02,0.10),U(0.02,0.10))$. \\
    \cmidrule[l=-1]{2-4}
    & \rotatebox[origin=c]{90}{Gravity} &
    \textit{Direction error} & Compose attitude with an OU rotation residual with correlation time $2.0\,\mathrm{s}$ and per-axis driving scale $\sigma_{\mathrm{eff}}=\sigma_0 c[1+d\min(\|\widetilde{\omega}\|/(1\,\mathrm{rad/s})+\|a\|/g,1)]$, where $\sigma_0\sim U(0.4,1.2)^\circ$, $d\sim U(0.3,1.6)$, $\widetilde{\omega}$ is the simulated gyroscope measurement, and $a$ includes vibration; $c\sim U(1,1.7)$ for ground, $U(1,1.4)$ for handheld, and $c=1$ otherwise. \\
    \bottomrule
  \end{tblr}%
  }
  \par\vspace{3pt}
  \begin{minipage}{\columnwidth}
    \raggedright
    \textit{Notes:} $U(a,b)$ denotes uniform distribution, $\mathcal{N}(\mu,\sigma^2)$ denotes Gaussian distribution, and OU denotes Ornstein-Uhlenbeck.
    The temporal persistent processes use entry probability $\lambda\Delta t$ and recovery probability $\Delta t/\tau$ per update, where $\Delta t$ is the update interval and $\tau$ is the mean active duration.
  \end{minipage}
\end{table}

\section{Experiments}
\label{sec:experiments}

\subsection{Experiment Setup}
\label{sec:setup}

\subsubsection{Datasets and Protocol}

The simulated training data is described in \autoref{sec:sim2real}.
The real test data comprises 24 recordings from CREPES-X \cite{li2025crepesx} and CT-RIO \cite{lu2026ctrio}, evenly divided among six scenarios: Line-Of-Sight (\textbf{\texttt{LOS}}), Non-Line-Of-Sight (\textbf{\texttt{NLOS}}), High Dynamic Motion (\textbf{\texttt{HDM}}), Shift Dynamic Motion (\textbf{\texttt{SDM}}), Multi-Robot Platform (\textbf{\texttt{MRP}}), and TEN-device heterogeneous (\textbf{\texttt{TEN}}).
We follow the CREPES-X \cite{li2025crepesx} and CT-RIO \cite{lu2026ctrio} protocols and compute position and rotation RMSE in Euclidean distance and geodesic angle.
For each of three random seeds, frames 1–150 are used to fine-tune AnyAmber \cite{wang2026anyamber}, frames 151–200 are used for initialization, and frames 201 onward are used for evaluation in each sequence.
Evaluation is performed on the same set of valid reference-to-robot samples, with robot $0$ serving as the reference robot.
Each scenario-level result is the mean of the four corresponding sequence-level RMSE values, whereas \textit{overall} is the mean of the sequence-level RMSE values over all 24 sequences.
We mark a sequence, scenario, or overall result as \textit{failed} when its position RMSE exceeds $100.0\,\mathrm{cm}$.

\subsubsection{Implementation Details}

The estimator operates on a shared $50\,\mathrm{Hz}$ timeline defined by the camera frames.
At each frame timestamp, the most recent preceding UWB measurement is used, while the IMU measurements acquired since the preceding frame are uniformly resampled to four samples to form the inertial window.
The network uses four recurrent factor graph layers, 128-dimensional node states, 64-dimensional factor states, four attention heads, and 256-dimensional feed-forward blocks, totaling 1.67 million parameters.
Training uses AdamW \cite{loshchilov2019decoupled} with batch size $25$, weight decay $10^{-4}$, an initial learning rate of $10^{-3}$, 10 warm-up epochs, cosine decay to $10^{-5}$, 250-frame chunks, and unit-norm gradient clipping.
Each model is trained for $1000$ epochs with $100$ steps per epoch, and the final checkpoint is used for evaluation.
Training requires approximately $8.9$ hours on a single NVIDIA GeForce RTX 3090Ti GPU.

\subsection{Comparison with Existing Methods}
\label{sec:comparison}

\begin{figure*}[t]
  \centering
  \includegraphics[width=.98\textwidth]{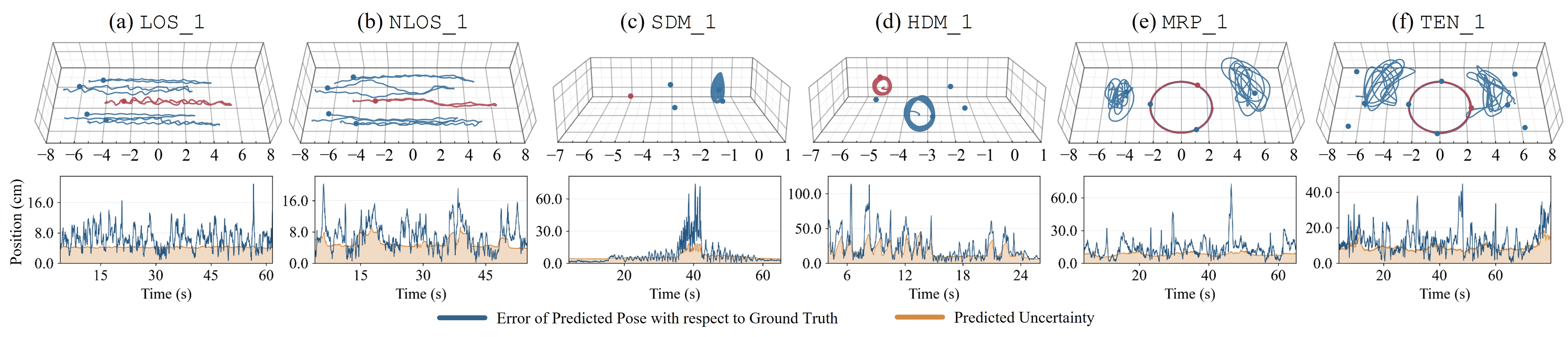}
  \caption{
    Visualization of \texttt{LOS\_1}, \texttt{NLOS\_1}, \texttt{SDM\_1}, \texttt{HDM\_1}, \texttt{MRP\_1}, and \texttt{TEN\_1}.
    The top row shows the ground-truth trajectories of the reference robot (red) and other robots (blue).
    The bottom row shows the error of NeuRIO's predicted pose of device 1 with respect to the ground-truth (blue) and the predicted uncertainty (orange).
    The uncertainty is visualized as the predicted position RMS scale $\sqrt{3\exp(s_p)}$, where $s_p$ is the predicted position log variance.
  }
  \label{fig:trajectories}
\end{figure*}

We compare NeuRIO with existing model-based and learned methods on all 24 real-world collected test sequences, and the results are summarized in \autoref{tab:main_results}.
CREPES \cite{xun2023crepes}, CREPES-X \cite{li2025crepesx}, and CT-RIO \cite{lu2026ctrio} are included
as filter/optimization baselines.
AnyAmber \cite{wang2026anyamber} is included as the most relevant published learned comparator on the overlapping relative-localization setting, although its full task also covers anonymous association, varying numbers of UWB devices per robot, and the presence or absence of bearing measurements.
For each sequence, the full pipeline of AnyAmber is evaluated in two configurations: pretrained by Wang et al. \cite{wang2026anyamber} and fine-tuned for three epochs on the first 150 frames of the target sequence.
NeuRIO remains fully frozen and uses no target trajectory for training or adaptation.

As shown in \autoref{tab:main_results}, NeuRIO achieves an overall RMSE of $14.1\,\mathrm{cm}$ / $3.9^\circ$ across the 24 real-world sequences.
AnyAmber is substantially less robust under recurrent inference.
Using its own previous estimates as pose priors, all pretrained sequences exceed the $100\,\mathrm{cm}$ position failure threshold.
Fine-tuning improves performance in the \textbf{\texttt{LOS}}, \textbf{\texttt{NLOS}}, and \textbf{\texttt{SDM}} scenarios, but RMSEs of all sequences in \textbf{\texttt{HDM}}, \textbf{\texttt{MRP}}, and \textbf{\texttt{TEN}} still exceed the $100\,\mathrm{cm}$ threshold.
Providing AnyAmber with ground-truth pose priors leads to considerably stronger results.
Under this protocol, three-epoch fine-tuning reduces its overall error from $30.4\,\mathrm{cm}$ / $5.0^\circ$ to $22.3\,\mathrm{cm}$ / $4.0^\circ$.
Nevertheless, the frozen NeuRIO remains more accurate in both position and rotation, despite requiring neither ground-truth pose priors nor target-sequence fine-tuning.
\autoref{fig:trajectories} visualizes NeuRIO's outputs of one sequence in each scenario.
The results show that NeuRIO can directly predict relative poses under tested real-world scenarios, and the predicted uncertainty can reflect the actual error distribution although its magnitude is underestimated in some cases.

The optimization-based methods achieve the highest geometric accuracy.
CREPES-X and CT-RIO obtain $9.9\,\mathrm{cm}$ / $2.4^\circ$ and $7.0\,\mathrm{cm}$ / $2.2^\circ$, respectively, while CREPES obtains $18.0\,\mathrm{cm}$ / $3.5^\circ$.
Although CREPES-X and CT-RIO are more accurate than NeuRIO, they require iterative optimization at each frame, which is computationally expensive and scales poorly with the number of robots.
This limits their applicability to large swarms and will be discussed next.

\begin{table}[t]
  \centering
  \caption{
    Comparison over 24 real-world sequences (RMSE in cm / deg).
  }
  \label{tab:main_results}
  \resizebox{\columnwidth}{!}{%
    \setlength{\tabcolsep}{2pt}
    \begin{tabular}{@{}l*{8}{c}@{}}
      \toprule
      \multirow{2}{*}{\vspace{-12pt}{Seq.}}
                       & \multirow{2}{*}{\vspace{-8pt}\MethodHeader{CREPES}}
                       & \multirow{2}{*}{\vspace{-8pt}\MethodHeader{CREPES-X}}
                       & \multirow{2}{*}{\vspace{-8pt}\MethodHeader{CT-RIO}}
                       & \multicolumn{4}{c}{\shortstack{AnyAmber}}
                       & \multirow{2}{*}{\vspace{-8pt}\MethodHeader{NeuRIO}}
      \\
      \cmidrule{5-8}
                       &                                                       &                         &                         & \MethodHeader{Pretrained} & \MethodHeader{Fine-tuned} & \MethodHeader{Pretrained                                                                  \\(GT prior)} & \MethodHeader{Fine-tuned\\(GT prior)} &
      \\
      \midrule
      \texttt{LOS\_1}  & \AlignedPair{6.4}{2.1}                                & \AlignedPair{6.3}{1.9}  & \AlignedPair{5.7}{1.9}  & failed                    & \AlignedPair{12.8}{4.1}   & \AlignedPair[2.3em]{9.3}{2.6}                                                                           & \AlignedPair[2.3em]{9.5}{2.8}         & \AlignedPair{7.1}{2.0}  \\
      \texttt{LOS\_2}  & \AlignedPair{8.6}{2.4}                                & \AlignedPair{7.8}{2.1}  & \AlignedPair{6.7}{2.0}  & failed                    & \AlignedPair{10.6}{3.0}   & \AlignedPair[2.3em]{9.6}{2.6}                                                                           & \AlignedPair[2.3em]{9.0}{2.2}         & \AlignedPair{8.6}{2.3}  \\
      \texttt{LOS\_3}  & \AlignedPair{8.5}{2.8}                                & \AlignedPair{6.9}{2.1}  & \AlignedPair{6.1}{2.1}  & failed                    & \AlignedPair{8.7}{3.0}    & \AlignedPair[2.3em]{9.6}{2.7}                                                                           & \AlignedPair[2.3em]{7.3}{2.0}         & \AlignedPair{7.8}{2.2}  \\
      \texttt{LOS\_4}  & \AlignedPair{7.2}{2.1}                                & \AlignedPair{8.0}{2.5}  & \AlignedPair{7.0}{2.5}  & failed                    & \AlignedPair{13.8}{4.4}   & \AlignedPair[2.3em]{9.9}{2.9}                                                                           & \AlignedPair[2.3em]{10.6}{2.9}        & \AlignedPair{8.5}{2.8}  \\
      \midrule
      \texttt{NLOS\_1} & \AlignedPair{6.5}{2.2}                                & \AlignedPair{5.0}{1.8}  & \AlignedPair{4.6}{1.8}  & failed                    & \AlignedPair{13.9}{69.7}  & \AlignedPair[2.3em]{9.7}{2.3}                                                                           & \AlignedPair[2.3em]{7.0}{1.8}         & \AlignedPair{6.7}{2.5}  \\
      \texttt{NLOS\_2} & \AlignedPair{8.4}{2.4}                                & \AlignedPair{6.7}{2.0}  & \AlignedPair{5.6}{2.0}  & failed                    & \AlignedPair{23.8}{71.4}  & \AlignedPair[2.3em]{10.1}{2.9}                                                                          & \AlignedPair[2.3em]{8.4}{2.0}         & \AlignedPair{8.0}{2.3}  \\
      \texttt{NLOS\_3} & \AlignedPair{6.8}{2.3}                                & \AlignedPair{5.9}{2.0}  & \AlignedPair{5.4}{2.0}  & failed                    & \AlignedPair{29.4}{92.3}  & \AlignedPair[2.3em]{10.1}{2.5}                                                                          & \AlignedPair[2.3em]{7.8}{1.8}         & \AlignedPair{6.9}{2.2}  \\
      \texttt{NLOS\_4} & \AlignedPair{8.0}{2.5}                                & \AlignedPair{7.0}{2.1}  & \AlignedPair{6.1}{2.0}  & failed                    & \AlignedPair{16.7}{31.2}  & \AlignedPair[2.3em]{11.2}{2.7}                                                                          & \AlignedPair[2.3em]{8.7}{2.1}         & \AlignedPair{8.2}{2.2}  \\
      \midrule
      \texttt{SDM\_1}  & \AlignedPair{4.7}{3.5}                                & \AlignedPair{4.2}{1.6}  & \AlignedPair{2.8}{1.3}  & failed                    & \AlignedPair{10.6}{18.6}  & \AlignedPair[2.3em]{9.7}{2.5}                                                                           & \AlignedPair[2.3em]{6.7}{1.8}         & \AlignedPair{9.3}{8.1}  \\
      \texttt{SDM\_2}  & \AlignedPair{3.8}{2.2}                                & \AlignedPair{3.7}{1.4}  & \AlignedPair{2.5}{1.1}  & failed                    & \AlignedPair{9.3}{60.6}   & \AlignedPair[2.3em]{9.7}{2.4}                                                                           & \AlignedPair[2.3em]{6.0}{1.6}         & \AlignedPair{4.9}{1.9}  \\
      \texttt{SDM\_3}  & \AlignedPair{5.8}{2.6}                                & \AlignedPair{4.1}{1.8}  & \AlignedPair{2.6}{1.5}  & failed                    & \AlignedPair{13.8}{57.9}  & \AlignedPair[2.3em]{10.6}{2.1}                                                                          & \AlignedPair[2.3em]{10.9}{1.7}        & \AlignedPair{9.2}{8.0}  \\
      \texttt{SDM\_4}  & \AlignedPair{3.5}{1.4}                                & \AlignedPair{4.1}{1.3}  & \AlignedPair{2.1}{1.1}  & failed                    & \AlignedPair{9.4}{56.6}   & \AlignedPair[2.3em]{11.1}{2.5}                                                                          & \AlignedPair[2.3em]{5.8}{1.7}         & \AlignedPair{4.7}{2.1}  \\
      \midrule
      \texttt{HDM\_1}  & \AlignedPair{21.1}{4.1}                               & \AlignedPair{19.6}{3.7} & \AlignedPair{14.7}{2.8} & failed                    & failed                    & \AlignedPair[2.3em]{25.6}{5.1}                                                                          & \AlignedPair[2.3em]{25.8}{4.8}        & \AlignedPair{33.2}{4.9} \\
      \texttt{HDM\_2}  & \AlignedPair{16.3}{2.7}                               & \AlignedPair{14.0}{2.7} & \AlignedPair{10.6}{2.2} & failed                    & failed                    & \AlignedPair[2.3em]{24.4}{4.0}                                                                          & \AlignedPair[2.3em]{18.6}{2.9}        & \AlignedPair{21.8}{3.0} \\
      \texttt{HDM\_3}  & \AlignedPair{26.2}{3.6}                               & \AlignedPair{17.3}{2.3} & \AlignedPair{7.5}{1.8}  & failed                    & failed                    & \AlignedPair[2.3em]{21.6}{3.8}                                                                          & \AlignedPair[2.3em]{16.6}{3.0}        & \AlignedPair{14.7}{3.0} \\
      \texttt{HDM\_4}  & \AlignedPair{19.7}{4.7}                               & \AlignedPair{14.4}{2.8} & \AlignedPair{9.9}{2.2}  & failed                    & failed                    & \AlignedPair[2.3em]{26.0}{4.7}                                                                          & \AlignedPair[2.3em]{24.6}{4.2}        & \AlignedPair{29.8}{5.1} \\
      \midrule
      \texttt{MRP\_1}  & \AlignedPair{18.2}{4.8}                               & \AlignedPair{10.7}{3.6} & \AlignedPair{8.1}{3.6}  & failed                    & failed                    & \AlignedPair[2.3em]{49.3}{9.0}                                                                          & \AlignedPair[2.3em]{29.8}{7.0}        & \AlignedPair{12.4}{5.3} \\
      \texttt{MRP\_2}  & \AlignedPair{19.3}{4.6}                               & \AlignedPair{10.5}{3.3} & \AlignedPair{8.4}{3.2}  & failed                    & failed                    & \AlignedPair[2.3em]{47.2}{8.6}                                                                          & \AlignedPair[2.3em]{33.6}{7.0}        & \AlignedPair{12.5}{4.7} \\
      \texttt{MRP\_3}  & \AlignedPair{39.2}{4.6}                               & \AlignedPair{16.3}{3.1} & \AlignedPair{14.9}{2.9} & failed                    & failed                    & \AlignedPair[2.3em]{56.9}{9.6}                                                                          & \AlignedPair[2.3em]{31.0}{7.1}        & \AlignedPair{35.7}{7.2} \\
      \texttt{MRP\_4}  & \AlignedPair{42.0}{5.0}                               & \AlignedPair{16.1}{3.2} & \AlignedPair{13.4}{2.9} & failed                    & failed                    & \AlignedPair[2.3em]{58.4}{9.3}                                                                          & \AlignedPair[2.3em]{34.7}{7.7}        & \AlignedPair{29.8}{6.2} \\
      \midrule
      \texttt{TEN\_1}  & \AlignedPair{13.7}{2.7}                               & \AlignedPair{12.5}{2.2} & \AlignedPair{5.1}{2.0}  & failed                    & failed                    & \AlignedPair[2.3em]{51.4}{5.8}                                                                          & \AlignedPair[2.3em]{27.7}{3.7}        & \AlignedPair{13.3}{2.3} \\
      \texttt{TEN\_2}  & \AlignedPair{14.4}{2.5}                               & \AlignedPair{11.9}{2.1} & \AlignedPair{5.8}{2.0}  & failed                    & failed                    & \AlignedPair[2.3em]{48.7}{5.6}                                                                          & \AlignedPair[2.3em]{22.5}{3.1}        & \AlignedPair{15.0}{2.3} \\
      \texttt{TEN\_3}  & \AlignedPair{65.0}{8.5}                               & \AlignedPair{12.9}{2.7} & \AlignedPair{5.6}{2.7}  & failed                    & failed                    & failed                                                                                                  & failed                                & \AlignedPair{14.6}{5.6} \\
      \texttt{TEN\_4}  & \AlignedPair{58.5}{8.0}                               & \AlignedPair{12.5}{2.7} & \AlignedPair{5.8}{2.2}  & failed                    & failed                    & \AlignedPair[2.3em]{59.9}{5.8}                                                                          & \AlignedPair[2.3em]{44.0}{4.2}        & \AlignedPair{15.3}{4.7} \\
      \midrule
      Overall          & \AlignedPair{18.0}{3.5}                               & \AlignedPair{9.9}{2.4}  & \AlignedPair{7.0}{2.2}  & failed                    & failed                    & \AlignedPair[2.3em]{30.4}{5.0}                                                                          & \AlignedPair[2.3em]{22.3}{4.0}        & \AlignedPair{14.1}{3.9} \\
      \bottomrule
    \end{tabular}
  }
\end{table}

\subsection{Accuracy and Scalability}
\label{sec:accuracy_scalability}

\begin{figure}[t]
  \centering
  \includegraphics[width=.98\columnwidth]{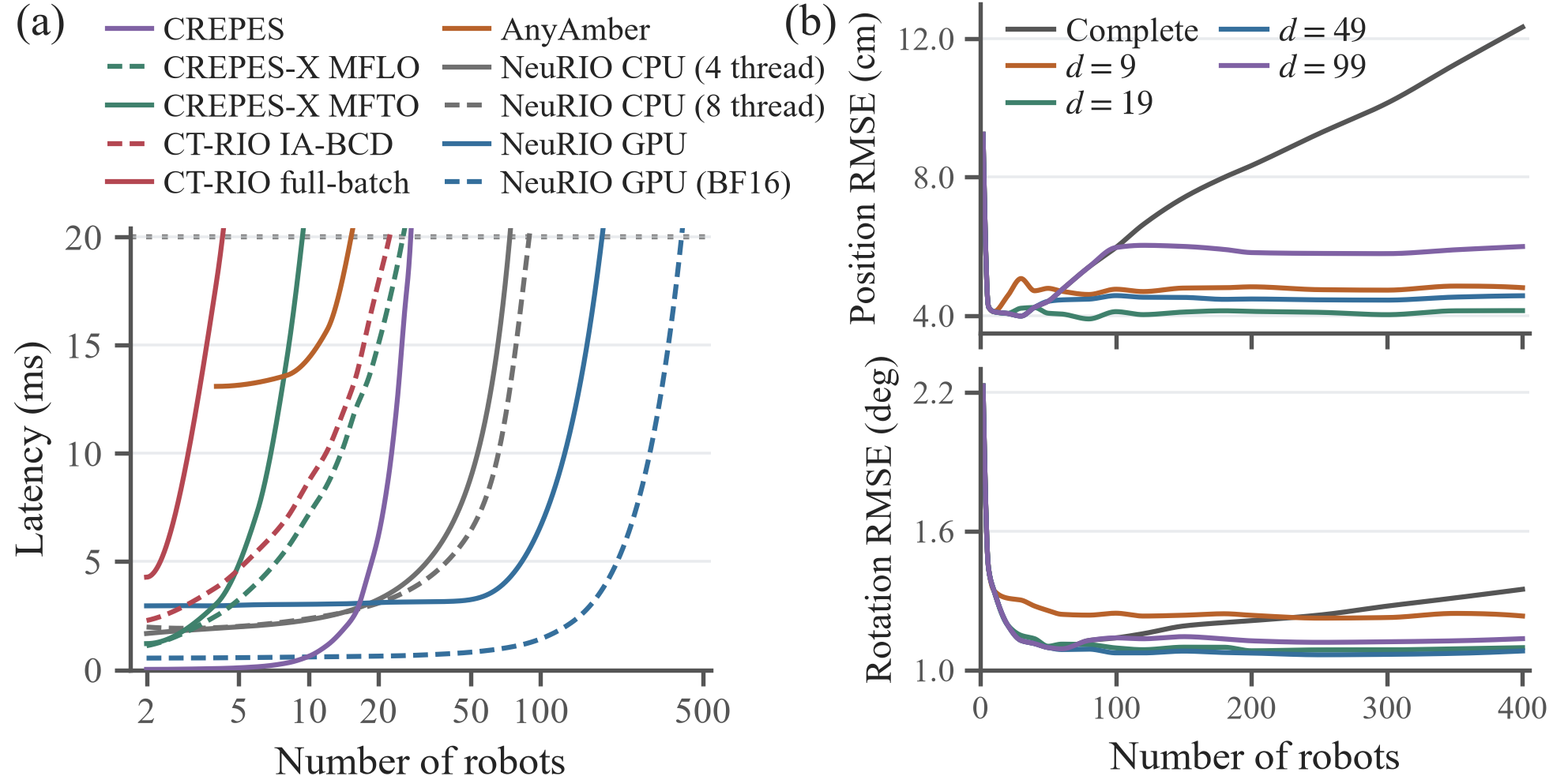}
  \caption{
    Computational scalability and accuracy in simulation as the number of robots increases.
    (a) Mean computation time, with a $20\,\mathrm{ms}$ reference corresponding to $50\,\mathrm{Hz}$ sensing.
    AnyAmber uses published timings with MatchNet excluded (which is used for anonymous bearing association) \cite{wang2026anyamber}.
    (b) Position and rotation RMSE of the frozen network in simulation.
    Complete denotes a fully connected observation graph; otherwise, each target robot has $\min(d,n-1)$ observed neighbors.
  }
  \label{fig:runtime_scalability}
\end{figure}

\autoref{fig:runtime_scalability} evaluates the computational scalability of the compared estimators and the generalization of NeuRIO to larger team sizes and different observation densities.
For runtime, all methods are evaluated in simulation and on the same workstation equipped with an Intel Core i9-14900KF and an NVIDIA RTX 4070Ti SUPER.
Using $20\,\mathrm{ms}$ per update as the real-time reference, CREPES, CREPES-X MFLO, CREPES-X MFTO, CT-RIO IA-BCD, and CT-RIO full-batch remain below this threshold up to $27$, $24$, $8$, $22$, and $4$ robots, respectively.
In comparison, NeuRIO scales to $72$ robots with four CPU threads, $91$ robots with eight CPU threads, and $180$ robots with GPU FP32.
With GPU BF16, it remains within the same computation budget at $400$ robots.
When using BF16, no increase in RMSE is observed in the real-world sequences.
These results highlight the complementary operating regimes of the methods: CREPES-X and CT-RIO provide higher geometric accuracy, whereas NeuRIO supports substantially larger swarms at real-time rates.

We further evaluate the frozen NeuRIO model on synthetic teams ranging from $2$ to $400$ robots, as real-world experiments with hundreds of robots are impractical.
Under complete observation graphs, position RMSE grows from $4.1\,\mathrm{cm}$ at $10$ robots to $6.0\,\mathrm{cm}$ at $100$ robots and $12.3\,\mathrm{cm}$ at $400$ robots, while rotation RMSE remains near $1.1^\circ$ to $1.3^\circ$.
This degradation is substantially alleviated when the observation degree, namely the number of neighbors observed by the robot, is bounded.
With degree $d=49$, position RMSE remains approximately $4.4$ to $4.6\,\mathrm{cm}$ from $100$ to $400$ robots, with rotation RMSE around $1.1^\circ$.
These results suggest that the degradation at very large team sizes is primarily associated with the rapidly growing observation neighborhoods in dense graphs rather than the number of robots alone.
One possible explanation is that complete graphs require each robot to aggregate information from an increasingly large number of factors, producing neighborhood sizes far beyond those encountered during training and potentially diluting the contribution of informative observations.

\subsection{Training Team Size Generalization}
\label{sec:team_generalization}

We vary the maximum team size used during training and evaluate each resulting model on the full real test set.
The default training distribution contains three to ten robots, while \textbf{\texttt{TEN}} consists of real ten-robot sequences.
As shown in \autoref{tab:team_scaling}, training with at most three robots generalizes poorly to larger teams, yielding $35.9\,\mathrm{cm}$ / $8.2^\circ$ overall and $58.4\,\mathrm{cm}$ / $10.0^\circ$ on \textbf{\texttt{TEN}}.
Increasing the training limit to five robots substantially reduces the corresponding errors to $16.3\,\mathrm{cm}$ / $4.4^\circ$ and $18.1\,\mathrm{cm}$ / $4.7^\circ$.
With at most seven training robots, performance nearly saturates, reaching $14.0\,\mathrm{cm}$ / $3.9^\circ$ overall and $14.6\,\mathrm{cm}$ / $3.9^\circ$ on \textbf{\texttt{TEN}}, closely matching the default ten-robot training distribution.
These results show that NeuRIO transfers to unseen larger team sizes without architectural or parameter changes, and that exposure to moderate team sizes during training is sufficient to capture most of this generalization ability.

\begin{table}[ht]
  \centering
  \caption{
    Generalization across training numbers (RMSE in cm / deg).
  }
  \label{tab:team_scaling}
  \resizebox{\columnwidth}{!}{%
    \setlength{\tabcolsep}{3pt}
    \begin{tabular}{@{}lccccccc@{}}
      \toprule
      \shortstack{Max number} & \textbf{\texttt{LOS}}   & \textbf{\texttt{NLOS}}  & \textbf{\texttt{SDM}}   & \textbf{\texttt{HDM}}   & \textbf{\texttt{MRP}}    & \textbf{\texttt{TEN}}    & \textbf{Overall}        \\
      \midrule
      3 robots                & \AlignedPair{16.6}{3.6} & \AlignedPair{26.4}{6.2} & \AlignedPair{10.5}{4.7} & \AlignedPair{45.9}{7.0} & \AlignedPair{57.6}{17.5} & \AlignedPair{58.4}{10.0} & \AlignedPair{35.9}{8.2} \\
      5 robots                & \AlignedPair{9.5}{2.4}  & \AlignedPair{8.9}{2.6}  & \AlignedPair{6.5}{6.3}  & \AlignedPair{28.1}{4.1} & \AlignedPair{26.9}{6.4}  & \AlignedPair{18.1}{4.7}  & \AlignedPair{16.3}{4.4} \\
      7 robots                & \AlignedPair{8.6}{2.4}  & \AlignedPair{7.8}{2.4}  & \AlignedPair{6.1}{5.3}  & \AlignedPair{25.1}{4.1} & \AlignedPair{22.0}{5.5}  & \AlignedPair{14.6}{3.9}  & \AlignedPair{14.0}{3.9} \\
      10 robots               & \AlignedPair{8.0}{2.3}  & \AlignedPair{7.5}{2.3}  & \AlignedPair{7.0}{5.0}  & \AlignedPair{24.9}{4.0} & \AlignedPair{22.6}{5.8}  & \AlignedPair{14.6}{3.7}  & \AlignedPair{14.1}{3.9} \\
      \bottomrule
    \end{tabular}
  }
\end{table}

\subsection{Network Spatial Block Comparison}

We compare the proposed spatial block with GATv2 \cite{brody2022attentive}, GINE \cite{hu2020strategies}, GENConv \cite{li2020deepergcn}, TransformerConv \cite{shi2021masked}, ResGated \cite{bresson2017residual}, TokenGT \cite{kim2022tokengt}, and FGNN \cite{zhang2020factor} under the same frontend, node and directed-edge attributes, temporal components, and training protocol.
Only the within-frame spatial operator is replaced.
The former five use the PyTorch Geometric \cite{fey2019fast} implementations, while TokenGT \cite{kim2022tokengt} and FGNN \cite{zhang2020factor} are implemented in-house.

As shown in \autoref{tab:spatial_baselines}, NeuRIO achieves the best accuracy with RMSE of $14.1\,\mathrm{cm}$ / $3.9^\circ$.
FGNN is the closest alternative with RMSE of $18.3\,\mathrm{cm}$ / $5.9^\circ$, while the remaining baselines show larger degradation, particularly in the more challenging \textbf{\texttt{HDM}}, \textbf{\texttt{MRP}}, and \textbf{\texttt{TEN}} scenarios.
Across individual scenarios, NeuRIO obtains the lowest position RMSE in five scenarios and the lowest rotation RMSE in all six scenarios.
These results indicate that the proposed factor-graph with cross-attention message passing is better suited to the directed mutual observation structure than the compared generic graph operators and original FGNN formulation.

\begin{table}[ht]
  \centering
  \caption{
    Comparison of spatial blocks (RMSE in cm / deg).
  }
  \label{tab:spatial_baselines}
  \resizebox{\columnwidth}{!}{%
    \setlength{\tabcolsep}{3pt}
    \begin{tabular}{@{}lccccccc@{}}
      \toprule
      Spatial block   & \textbf{\texttt{LOS}}   & \textbf{\texttt{NLOS}}  & \textbf{\texttt{SDM}}   & \textbf{\texttt{HDM}}   & \textbf{\texttt{MRP}}    & \textbf{\texttt{TEN}}   & \textbf{Overall}        \\
      \midrule
      GATv2           & \AlignedPair{21.3}{4.4} & \AlignedPair{24.9}{5.0} & \AlignedPair{23.3}{5.6} & \AlignedPair{67.0}{9.1} & failed                   & \AlignedPair{87.4}{8.3} & \AlignedPair{56.3}{8.3} \\
      GINE            & \AlignedPair{12.4}{2.9} & \AlignedPair{13.4}{3.3} & \AlignedPair{13.5}{6.5} & \AlignedPair{47.7}{7.2} & \AlignedPair{44.9}{8.9}  & \AlignedPair{50.1}{6.5} & \AlignedPair{30.3}{5.9} \\
      GENConv         & \AlignedPair{12.3}{2.9} & \AlignedPair{12.1}{3.0} & \AlignedPair{14.2}{7.0} & \AlignedPair{48.9}{7.6} & \AlignedPair{42.2}{9.1}  & \AlignedPair{45.2}{6.0} & \AlignedPair{29.1}{5.9} \\
      TransformerConv & \AlignedPair{10.9}{2.6} & \AlignedPair{10.8}{2.8} & \AlignedPair{12.2}{5.9} & \AlignedPair{41.2}{5.9} & \AlignedPair{39.6}{8.1}  & \AlignedPair{39.5}{5.4} & \AlignedPair{25.7}{5.1} \\
      ResGated        & \AlignedPair{10.7}{2.7} & \AlignedPair{9.8}{2.7}  & \AlignedPair{10.9}{7.4} & \AlignedPair{41.3}{6.0} & \AlignedPair{37.0}{7.8}  & \AlignedPair{38.7}{5.3} & \AlignedPair{24.7}{5.3} \\
      TokenGT         & \AlignedPair{11.6}{2.4} & \AlignedPair{12.0}{3.0} & \AlignedPair{7.9}{7.6}  & \AlignedPair{38.3}{5.7} & \AlignedPair{37.3}{12.5} & \AlignedPair{42.3}{6.9} & \AlignedPair{24.9}{6.3} \\
      FGNN            & \AlignedPair{8.8}{3.7}  & \AlignedPair{8.7}{3.7}  & \AlignedPair{6.8}{5.7}  & \AlignedPair{25.6}{5.4} & \AlignedPair{38.2}{11.4} & \AlignedPair{21.6}{5.7} & \AlignedPair{18.3}{5.9} \\
      NeuRIO          & \AlignedPair{8.0}{2.3}  & \AlignedPair{7.5}{2.3}  & \AlignedPair{7.0}{5.0}  & \AlignedPair{24.9}{4.0} & \AlignedPair{22.6}{5.8}  & \AlignedPair{14.6}{3.7} & \AlignedPair{14.1}{3.9} \\
      \bottomrule
    \end{tabular}
  }
\end{table}

\subsection{Network Backbone Design Ablation}

\autoref{tab:ablation} evaluates the main temporal and representation choices in the NeuRIO backbone.
The complete model achieves $14.1\,\mathrm{cm}$ / $3.9^\circ$ overall.
Cross-frame recurrence is the most critical component.
Removing all recurrent memory causes the estimator to fail overall and in several challenging scenarios.
Among the layer-wise memories, removing node memory increases the overall error to $17.1\,\mathrm{cm}$ / $5.3^\circ$, while removing factor memory yields $15.7\,\mathrm{cm}$ / $4.3^\circ$.
In contrast, removing either frontend memory produces only small and mixed changes, indicating that the recurrent states within the spatial stack provide the larger contribution.
The representation and training choices also affect performance.
Removing gravity canonicalization increases the overall error to $18.1\,\mathrm{cm}$ / $5.3^\circ$, while removing the learned IMU encoder yields $16.4\,\mathrm{cm}$ / $4.2^\circ$.
Reducing the training chunk from $250$ to $50$ frames further degrades performance to $17.7\,\mathrm{cm}$ / $4.6^\circ$.

\begin{table}[ht]
  \centering
  \caption{
    Backbone ablation results (RMSE in cm / deg).
  }
  \label{tab:ablation}
  \resizebox{\columnwidth}{!}{%
    \setlength{\tabcolsep}{3pt}
    \begin{tabular}{@{}lccccccc@{}}
      \toprule
      Variant & \textbf{\texttt{LOS}}    & \textbf{\texttt{NLOS}} & \textbf{\texttt{SDM}}    & \textbf{\texttt{HDM}}   & \textbf{\texttt{MRP}}    & \textbf{\texttt{TEN}}    & \textbf{Overall}        \\
      \midrule
      NeuRIO  & \AlignedPair{8.0}{2.3}   & \AlignedPair{7.5}{2.3} & \AlignedPair{7.0}{5.0}   & \AlignedPair{24.9}{4.0} & \AlignedPair{22.6}{5.8}  & \AlignedPair{14.6}{3.7}  & \AlignedPair{14.1}{3.9} \\
      w/o GC  & \AlignedPair{9.6}{3.4}   & \AlignedPair{9.3}{3.1} & \AlignedPair{6.9}{3.6}   & \AlignedPair{33.5}{5.0} & \AlignedPair{30.3}{12.4} & \AlignedPair{19.0}{4.1}  & \AlignedPair{18.1}{5.3} \\
      w/o IE  & \AlignedPair{9.0}{2.4}   & \AlignedPair{8.5}{2.6} & \AlignedPair{6.2}{3.3}   & \AlignedPair{24.5}{4.3} & \AlignedPair{27.2}{7.7}  & \AlignedPair{22.8}{4.9}  & \AlignedPair{16.4}{4.2} \\
      w/o FNM & \AlignedPair{7.6}{2.3}   & \AlignedPair{7.1}{2.4} & \AlignedPair{5.6}{7.2}   & \AlignedPair{27.2}{4.0} & \AlignedPair{22.6}{5.3}  & \AlignedPair{14.0}{3.6}  & \AlignedPair{14.0}{4.1} \\
      w/o FFM & \AlignedPair{8.6}{2.4}   & \AlignedPair{7.7}{2.4} & \AlignedPair{6.1}{4.7}   & \AlignedPair{24.8}{4.0} & \AlignedPair{22.2}{5.4}  & \AlignedPair{15.6}{4.2}  & \AlignedPair{14.2}{3.9} \\
      w/o LNM & \AlignedPair{9.0}{3.2}   & \AlignedPair{8.1}{3.0} & \AlignedPair{7.0}{4.9}   & \AlignedPair{27.3}{5.4} & \AlignedPair{29.4}{9.6}  & \AlignedPair{21.6}{5.7}  & \AlignedPair{17.1}{5.3} \\
      w/o LFM & \AlignedPair{8.0}{2.3}   & \AlignedPair{7.4}{2.4} & \AlignedPair{5.5}{6.0}   & \AlignedPair{36.6}{5.1} & \AlignedPair{22.5}{6.2}  & \AlignedPair{14.2}{3.5}  & \AlignedPair{15.7}{4.3} \\
      w/o AM  & \AlignedPair{29.8}{16.3} & failed                 & \AlignedPair{78.5}{31.9} & failed                  & failed                   & \AlignedPair{46.3}{20.7} & failed                  \\
      50-FTC  & \AlignedPair{8.3}{2.3}   & \AlignedPair{8.3}{2.5} & \AlignedPair{6.3}{5.5}   & \AlignedPair{41.4}{5.8} & \AlignedPair{26.0}{6.7}  & \AlignedPair{16.1}{4.5}  & \AlignedPair{17.7}{4.6} \\
      \bottomrule
    \end{tabular}
  }
  \par\vspace{3pt}
  \begin{minipage}{\columnwidth}
    \footnotesize\raggedright
    \textit{Abbreviations:} GC, gravity canonicalization; IE, IMU encoding; FNM, frontend node memory; FFM, frontend factor memory; LNM, layer node memory; LFM, layer factor memory; AM, all memory; FTC, frames per training chunk.
  \end{minipage}
\end{table}

\subsection{Simulator Feature Design Ablation}

\autoref{tab:simulator_ablation} examines the effect of each simulation component on the sim-to-real transfer.
Replacing persistent corruptions with frame-wise independent failures increases the overall error from $14.1\,\mathrm{cm}$ / $3.9^\circ$ to $23.7\,\mathrm{cm}$ / $6.3^\circ$.
Exposure to missing and corrupted observations is also important: removing camera corruption, camera dropout, or UWB dropout increases the overall position RMSE to $19.1\,\mathrm{cm}$, $20.4\,\mathrm{cm}$, and $20.2\,\mathrm{cm}$, respectively.
Camera corruption removal produces the largest overall rotation error of $6.6^\circ$.
The remaining components have smaller effects.
Removing IMU corruption, UWB corruption, or gravity errors causes only moderate changes.
Overall, the ablation indicates that the main transfer benefit comes from modeling failures as temporally persistent processes and exposing the model to realistic observation loss and outliers.

\begin{table}[ht]
  \centering
  \caption{
    Simulator ablation results (RMSE in cm / deg).
  }
  \label{tab:simulator_ablation}
  \resizebox{\columnwidth}{!}{%
    \setlength{\tabcolsep}{3pt}
    \begin{tabular}{@{}lccccccc@{}}
      \toprule
      Simulator & \textbf{\texttt{LOS}}  & \textbf{\texttt{NLOS}}  & \textbf{\texttt{SDM}}   & \textbf{\texttt{HDM}}   & \textbf{\texttt{MRP}}    & \textbf{\texttt{TEN}}   & \textbf{Overall}        \\
      \midrule
      Default   & \AlignedPair{8.0}{2.3} & \AlignedPair{7.5}{2.3}  & \AlignedPair{7.0}{5.0}  & \AlignedPair{24.9}{4.0} & \AlignedPair{22.6}{5.8}  & \AlignedPair{14.6}{3.7} & \AlignedPair{14.1}{3.9} \\
      w/o Temp. & \AlignedPair{8.7}{2.7} & \AlignedPair{9.6}{3.1}  & \AlignedPair{7.1}{7.1}  & \AlignedPair{41.9}{6.5} & \AlignedPair{40.5}{11.8} & \AlignedPair{34.4}{6.5} & \AlignedPair{23.7}{6.3} \\
      w/o CamD  & \AlignedPair{8.9}{2.4} & \AlignedPair{10.1}{3.0} & \AlignedPair{10.2}{8.3} & \AlignedPair{44.4}{6.3} & \AlignedPair{30.5}{7.3}  & \AlignedPair{18.3}{4.7} & \AlignedPair{20.4}{5.3} \\
      w/o CamC  & \AlignedPair{9.2}{3.0} & \AlignedPair{8.1}{3.3}  & \AlignedPair{6.1}{4.0}  & \AlignedPair{35.4}{5.7} & \AlignedPair{41.1}{20.0} & \AlignedPair{14.9}{3.8} & \AlignedPair{19.1}{6.6} \\
      w/o UWBD  & \AlignedPair{8.3}{2.3} & \AlignedPair{7.7}{2.5}  & \AlignedPair{5.8}{6.1}  & \AlignedPair{26.3}{4.2} & \AlignedPair{29.5}{5.6}  & \AlignedPair{43.7}{4.0} & \AlignedPair{20.2}{4.1} \\
      w/o UWBC  & \AlignedPair{8.5}{2.4} & \AlignedPair{7.4}{2.4}  & \AlignedPair{7.0}{5.6}  & \AlignedPair{24.8}{4.0} & \AlignedPair{22.7}{6.4}  & \AlignedPair{17.0}{3.8} & \AlignedPair{14.6}{4.1} \\
      w/o IMUC  & \AlignedPair{7.9}{2.3} & \AlignedPair{7.5}{2.3}  & \AlignedPair{5.2}{7.0}  & \AlignedPair{26.5}{4.1} & \AlignedPair{22.0}{5.4}  & \AlignedPair{16.2}{4.0} & \AlignedPair{14.2}{4.2} \\
      w/o GraE  & \AlignedPair{8.4}{2.6} & \AlignedPair{8.2}{2.7}  & \AlignedPair{6.5}{5.0}  & \AlignedPair{24.6}{4.5} & \AlignedPair{23.8}{6.1}  & \AlignedPair{19.9}{5.8} & \AlignedPair{15.2}{4.5} \\
      \bottomrule
    \end{tabular}
  }
  \par\vspace{3pt}
  \begin{minipage}{\columnwidth}
    \footnotesize\raggedright
    \textit{Abbreviations:} Temp., temporal persistence; CamD, camera dropout; CamC, camera corruption; UWBD, UWB dropout; UWBC, UWB corruption; IMUC, IMU corruption; GraE, Gravity errors.
  \end{minipage}
\end{table}

\section{Conclusion}
\label{sec:conclusion}

We presented NeuRIO, a streaming neural estimator for multi-robot relative localization from bearing, range, and inertial measurements.
By combining gravity-aligned representations with recurrent robot and factor states, NeuRIO captures spatial interactions and temporal context with a shared model that generalizes across team sizes.
Trained entirely on simulated data, the frozen model achieves $14.1\,\mathrm{cm}$ / $3.9^\circ$ overall RMSE across $24$ real-world sequences without real-world training or adaptation.
The experiments further demonstrate scalable inference and show that recurrent memory, temporally persistent sensing failures, and exposure to missing and corrupted observations are important for zero-shot transfer.
Future work can extend the estimator toward richer spatiotemporal modeling such as multi-frame processing with sliding-window, and equivariant representations that directly encode the underlying $\SEthree$ geometry.

\bibliographystyle{IEEEtran}
\bibliography{root}

\end{document}